\documentclass[sigconf,natbib=true,anonymous=false]{acmart}

\usepackage{amsmath,amsfonts,bm}

\def\eqref#1{equation~\ref{#1}}

\def\1{\bm{1}}

\DeclareMathAlphabet{\mathsfit}{\encodingdefault}{\sfdefault}{m}{sl}
\SetMathAlphabet{\mathsfit}{bold}{\encodingdefault}{\sfdefault}{bx}{n}

\usepackage{hyperref}
\usepackage{url}
\usepackage{wrapfig}
\usepackage{graphicx}
\usepackage{subcaption} 
\usepackage{multirow}
\usepackage{makecell}
\usepackage[table]{xcolor}
\usepackage{tabularx}
\definecolor{mygold}{rgb}{1, 0.925, 0.792}

\definecolor{darkgreen}{RGB}{0,160,0}

\usepackage{balance}

\title[CoRoVA: Compressed Representations for Vector-Augmented Code Completion]{CoRoVA: Compressed Representations \\for Vector-Augmented Code Completion}

\author{Daria Cherniuk}
\affiliation{%
  \institution{AXXX}
  \city{Moscow}
  \country{Russia}
}
\email{kamikazizen@gmail.com}

\author{Nikita Sukhorukov}
\affiliation{%
  \institution{AXXX}
  \city{Moscow}
  \country{Russia}
}

\author{Danil Gusak}
\affiliation{%
  \institution{AXXX}
  \city{Moscow}
  \country{Russia}
}
\email{gusak.contact@gmail.com}

\author{Nikita Sushko}
\affiliation{%
  \institution{AXXX}
  \city{Moscow}
  \country{Russia}
}
\email{gorakievskaya@gmail.com}

\author{Danil Sivtsov}
\affiliation{%
  \institution{AXXX}
  \city{Moscow}
  \country{Russia}
}
\email{sivtsovdt@gmail.com}

\author{Elena Tutubalina}
\affiliation{%
  \institution{AXXX}
  \city{Moscow}
  \country{Russia}
}
\email{tutubalinaev@gmail.com}

\author{Evgeny Frolov}
\affiliation{%
  \institution{AXXX\\HSE University}
  \city{Moscow}
  \country{Russia}
}
\email{evfro@live.ru}

\AtBeginDocument{%
  }

\acmDOI{XXXXXXX.XXXXXXX}
\copyrightyear{2026}
\acmYear{2026}
\setcopyright{acmlicensed}
\acmConference[SIGIR '26]{Proceedings of the 49th International ACM SIGIR Conference on Research and Development in Information Retrieval}{July 20--24, 2026}{Melbourne | Naarm, Australia}
\acmBooktitle{Proceedings of the 49th International ACM SIGIR Conference on Research and Development in Information Retrieval (SIGIR '26), July 20--24, 2026, Melbourne | Naarm, Australia}



\begin{abstract}

Retrieval-augmented generation has emerged as one of the most effective approaches for code completion enhancement, especially when repository-level context is important. However, adding this extra retrieved context significantly increases sequence length, raises prefill cost, and degrades time-to-first-token (TTFT), which slows down inference -- a critical limitation for interactive settings such as IDEs.
In this work, we introduce CoRoVA, a framework that compresses context into compact, semantically rich representations that remain interpretable to code LLMs. This improves generation quality while reducing prompt augmentation to only a few compressed single-token vectors.
Our approach requires training only a small projector module and introduces negligible additional latency, yet it significantly improves the prediction quality of code LLMs.
Our experiments show that CoRoVA enables a 20-38\% reduction in TTFT on completion tasks compared to uncompressed RAG.

\end{abstract}

\begin{teaserfigure}
\centering  \includegraphics[width=0.89\textwidth]{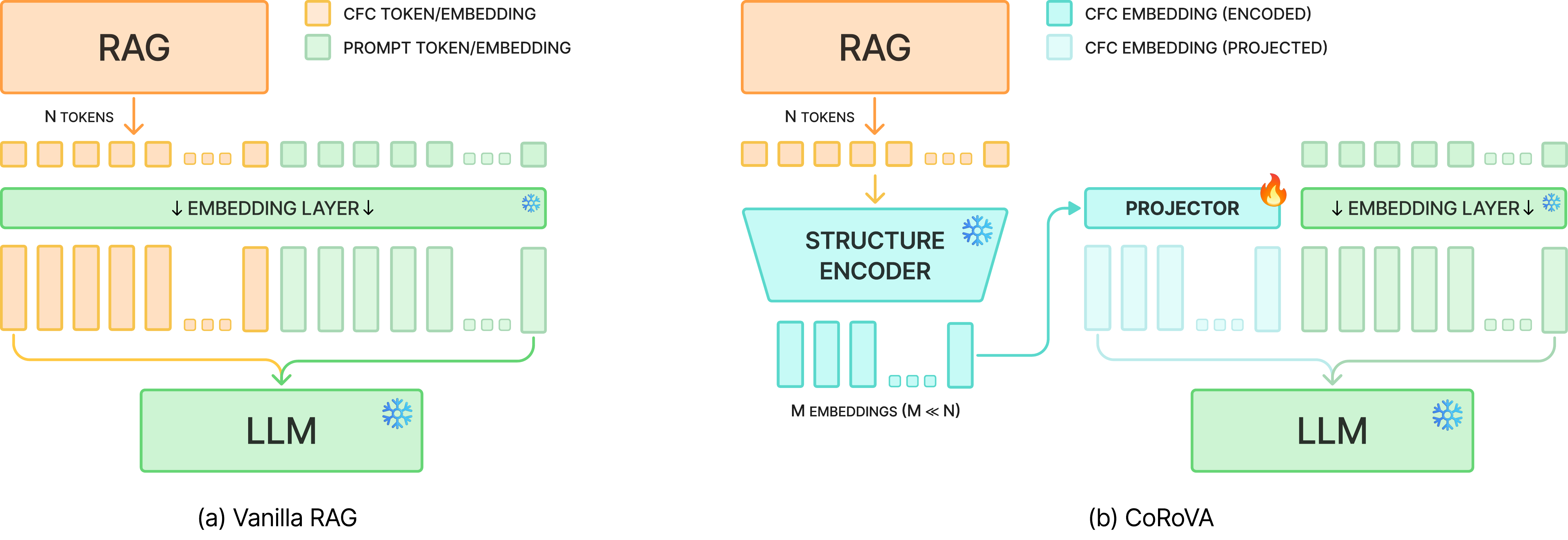}
 \caption{Comparison between Vanilla RAG~(a) and CoRoVA~(b) architectures. Instead of retrieving code passages and putting them into the context of the reader language model, CoRoVA uses a pretrained encoder to compress the code snippets and projects them into continuous tokens, thus reducing the final prompt processing time.}
  \label{fig:all}
\end{teaserfigure}

\begin{document}

\begin{abstract}

Retrieval-augmented generation has emerged as one of the most effective approaches for code completion enhancement, especially when repository-level context is important. However, adding this extra retrieved context significantly increases sequence length, raises prefill cost, and degrades time-to-first-token (TTFT), which slows down inference -- a critical limitation for interactive settings such as IDEs.
In this work, we introduce CoRoVA, a framework that compresses context into compact, semantically rich representations that remain interpretable to code LLMs. This improves generation quality while reducing prompt augmentation to only a few compressed single-token vectors.
Our approach requires training only a small projector module and introduces negligible additional latency, yet it significantly improves the prediction quality of code LLMs.
Our experiments show that CoRoVA enables a 20-38\% reduction in TTFT on completion tasks compared to uncompressed RAG.

\end{abstract}

\begin{abstract}

Retrieval-augmented generation has emerged as one of the most effective approaches for code completion enhancement, especially when repository-level context is important. However, adding this extra retrieved context significantly increases sequence length, raises prefill cost, and degrades time-to-first-token (TTFT), which slows down inference -- a critical limitation for interactive settings such as IDEs.
In this work, we introduce CoRoVA, a framework that compresses context into compact, semantically rich representations that remain interpretable to code LLMs. This improves generation quality while reducing prompt augmentation to only a few compressed single-token vectors.
Our approach requires training only a small projector module and introduces negligible additional latency, yet it significantly improves the prediction quality of code LLMs.
Our experiments show that CoRoVA enables a 20-38\% reduction in TTFT on completion tasks compared to uncompressed RAG.

\end{abstract}

\maketitle

\section{Introduction}
Modern code completion systems are increasingly deployed in interactive development environments (IDEs) as central tools. Code editors such as Windsurf\footnote{\href{https://windsurf.com/}{Windsurf homepage: https://windsurf.com/}} and Cursor\footnote{\href{https://cursor.com/}{Cursor homepage: https://cursor.com/}} started integrating large language models (LLMs) to provide single- and multi-line predictions, which can substantially improve developer productivity. However, these capabilities also come with strict latency requirements: even small delays in time-to-first-token (TTFT) break the interactive coding experience, making suggestion features unusable in practice. At the same time, high-quality completions often require \emph{repository-level context}: the correct line may depend on names, types, helper functions, and conventions defined elsewhere in the codebase rather than in the local window.

Retrieval-augmented generation (RAG)~\citep{lewis2021retrievalaugmentedgenerationknowledgeintensivenlp} naturally addresses the need for non-local context, and it has been widely adopted to improve both QA and completion quality, since it allows models to incorporate external context such as documentation, relevant snippets of code, or function declarations into the prompt (Fig.~\ref{fig:all}a). However, these additional retrieved tokens significantly increase prompt processing time and, consequently, TTFT, making vanilla RAG \emph{less practical for latency-critical settings such as code completion}.

A promising solution for this problem is context compression via embedding projection. Originally introduced in multimodal models such as Flamingo~\citep{alayrac2022flamingovisuallanguagemodel} and LLaVA~\citep{liu2023visualinstructiontuning}, where a dedicated vision encoder and a lightweight projector map high-dimensional image embeddings into a compact token sequence for the language model. Subsequent works (e.g., xRAG~\citep{cheng2024xragextremecontextcompression}) extended this idea to textual retrieval, showing that compressed representations can match vanilla RAG performance while dramatically reducing inference cost.

Despite this progress, no prior work has studied embedding-projection-style compression for the code completion task, where the latency-quality trade-off is particularly severe. 
Furthermore, existing training objectives (e.g., cross-entropy) are often \emph{misaligned with developer-relevant code-generation quality metrics} such as Exact Match (EM) and Edit Similarity (ES)~\cite{tan2024promptbasedcodecompletionmultiretrieval, repoformer}, limiting the effectiveness of current approaches. 
Additionally, code has multiple modalities beyond surface text, such as abstract syntax trees (ASTs), control flow graphs (CFGs), and data flow graphs (DFGs), which can be incorporated into the retrieved embeddings to \emph{enrich the representations} with structural and semantic information.


In this work, we address all these challenges: i) we introduce \textbf{Co}mpressed \textbf{R}epresentati\textbf{o}ns for \textbf{V}ector-\textbf{A}ugmented (CoRoVA) code completion, a projection mechanism that incorporates retrieved code snippets into the code-generating model’s input while adding only a few tokens to the prompt length; ii) we devise a novel training pipeline based on a three-component composite loss, consisting of cross-entropy, an RL-based term that directly optimizes EM and ES, and a novel cosine-alignment loss that preserves distinctions in the compressed representations; and iii) we incorporate syntactic and structural information from code representations into the projected embeddings.


In summary, our main contributions are:

\begin{itemize}
    \item The proposed \textbf{CoRoVA} approach is the first to apply embedding projection to code completion tasks \textit{without} embedder or LLM fine-tuning, achieving higher quality with negligible latency increase relative to the base model without RAG, while maintaining 20-38\% lower latency compared to full RAG;
    \item Prior projection-training methods -- whether based solely on cross-entropy or on cross-entropy combined with auxiliary losses -- proved insufficient for code completion. To address this, we designed a novel composite loss that integrates cross-entropy, an RL-inspired component, and a cosine-alignment term that preserves distinctions in the compressed representations;
    \item We study the incorporation of structural and syntactic code modalities, such as ASTs, to investigate whether these alternative signals can improve representation quality.
\end{itemize}

All the code and projector weights will be available\footnote{Code: \url{https://anonymous.4open.science/r/CoRoVA}\label{github}} under a permissive license.


\section{Related Work}
\label{sec:relatedwork}
\subsection{Coding LLMs}

StarCoder~\citep{li2023starcodersourceyou} introduced a family of code generation models, including larger LLMs optimized for code-centric dialogue and smaller ones tailored for code completion. Trained on The Stack dataset~\citep{kocetkov2022stack3tbpermissively}, these models achieved strong performance, surpassing most prior approaches on both code completion and instruction-following benchmarks.
The Qwen-2.5-Coder series~\citep{hui2024qwen25codertechnicalreport} represented another significant advancement in code-focused LLMs. Trained on a proprietary mixture of data, the models were released in sizes ranging from 0.5B to 32B parameters and were designed to support text completion, code chat, and fill-in-the-middle tasks.


\subsection{Context Compression Methods}

Although modern decoder-only transformers benefit from inference optimizations such as KV-caching~\citep{pope2022efficientlyscalingtransformerinference} and efficient attention implementations like GQA~\citep{ainslie2023gqatraininggeneralizedmultiquery}, time per-token inference latency still grows linearly with context size. Since Retrieval Augmented Generation~\citep{lewis2021retrievalaugmentedgenerationknowledgeintensivenlp} retrieves information from the knowledge base and puts it into the context of language models, this increases the context size that needs to be processed and subsequently increases end-to-end latency. Consequently, \emph{compressing retrieved context into compact, informative representations} -- without sacrificing semantic relevance -- is \emph{essential for deploying RAG in practice}.

In the paper xRAG~\citep{cheng2024xragextremecontextcompression} the authors propose an approach, which is similar to multi-modal language models training: they push the embedding vector of the retrieved text from textual encoder through a lightweight projector layer to align it with the reader model. The resulting architecture is trained in a two-stage manner. In the first stage, both the encoder and LLM are frozen, while the projection layer is trained with cross-entropy loss on paraphrases of the same document. During the second stage, the projector is trained on a mix of tasks such as reading comprehension, open-domain QA and summarization, adding self-distillation from RAG teacher via KL term along with usual negative log-likelihood loss. Models trained in this way perform competitively with vanilla RAG systems, while being much more efficient and having lower TTFT due to the reduction in prompt length.

Our approach is conceptually similar to xRAG method. By using a LLaVA-like projection from the encoder to the code completion model, we compress the retrieved context and maintain good generation quality, while lowering the TTFT. However, due to the specificity of our domain, we applied \emph{additional techniques} to increase code-specific metrics and quality of predictions. 
Furthermore, we train \emph{only the projector} with both the encoder and reader LLM frozen in a \emph{single stage manner}.

\subsection{Embedding Models for Code}

Many code-search embedding models are built by taking a pretrained decoder-only language model and fine-tuning it with contrastive learning to produce useful vector representations of code. One such model is Qwen3-Embedding-0.6B~\citep{zhang2025qwen3embeddingadvancingtext}, which is based on Qwen3-0.6B~\citep{yang2025qwen3technicalreport} LLM. Initialized from a powerful pretrained foundation, Qwen3-Embedding-0.6B shows competitive scores on MTEB~\citep{muennighoff2023mtebmassivetextembedding} benchmarks among similarly sized embedding models.

Moreover, some of the encoder models were trained not only on pure text and code data, but also on structured graphs, retrieved from code: data flow graphs (DFG) and abstract syntax trees (AST). Notable examples include GraphCodeBERT~\citep{guo2021graphcodebertpretrainingcoderepresentations} and UniXcoder~\citep{guo2022unixcoderunifiedcrossmodalpretraining}, which unified code, text and graph data to improve representation quality for code-understanding and retrieval tasks.

We evaluate two encoders utilizing different code representations to understand how the choice of representation affects both projection quality and inference latency (see Sections~\ref{sec:encoder_ablation} and~\ref{sec:rag_build_cost}).



\subsection{RL in Language Modeling}

Training language models solely for next-token prediction optimizes perplexity, but does not optimize other objectives such as safety, human preference alignment, or, specifically for our task, EM and ES scores.
In the Self-Critical Sequence Training (SCST) paper~\citep{rennie2017selfcriticalsequencetrainingimage}, a variation of REINFORCE~\citep{reinforce} with a baseline is applied to train an image captioning model. SCST uses the reward of the sequence produced by the current model under the test-time inference algorithm as the baseline, yielding an unbiased, lower-variance REINFORCE estimator.

Alternative reinforcement learning approaches have been proposed to further stabilize policy-gradient optimization. Proximal Policy Optimization (PPO)~\citep{schulman2017ppo} constrains policy updates through a learned value (critic) function and clipped surrogate objectives, but requires training additional models that are unnecessary in our setting with deterministic reward signals. More recently, Group Relative Policy Optimization (GRPO)~\citep{shao2024deepseekmath} eliminates the critic by normalizing rewards across multiple sampled rollouts per prompt, at the cost of computing gradients for each sampled trajectory. In contrast, SCST provides a computationally efficient solution by using a greedy rollout as a self-referential baseline, allowing stable optimization of sequence-level rewards without introducing extra learned components or incurring the overhead of multi-sample policy updates.

\section{Methodology}
\label{sec:methodology}
In this section, we present CoRoVA, a retrieval-augmented code completion approach that \textit{compresses retrieved repository context into a small set of continuous tokens}, enabling the code LLM to leverage rich cross-file evidence while avoiding the token overhead and latency of vanilla RAG.

\subsection{Model Architecture}

To reduce the number of retrieved tokens processed by the RAG reader model, CoRoVA compresses retrieved information into a compact, task-aware representation. Specifically, we use an off-the-shelf embedding model to encode \emph{each chunk} into a \emph{single dense vector}. This vector is then passed through a lightweight projector to align its semantic space with the reader LLM's token embeddings.

In practice, we take the top-10 retrieved chunks for each completion target and compress them into 10 projected contextual embeddings, which are concatenated with the prompt embeddings at the LLM input (Figure~\ref{fig:all}b). This design adds negligible latency (Section \ref{sec:practical} reports latency measurements), since we directly retrieve precomputed projections from the RAG database, without running the encoder or projector at the code completion time.

In our experiments, we use Qwen2.5-Coder and StarCoder2 model families as base code-completion LLMs. For encoding retrieved cross-file context, we adopt two embedding models: Qwen3-Embedding-0.6B~\citep{zhang2025qwen3embeddingadvancingtext} and UniXCoder~\citep{guo2022unixcoderunifiedcrossmodalpretraining}. 
The projector follows the LLaVA~\citep{liu2023visualinstructiontuning} projector architecture: a lightweight MLP with GELU~\citep{hendrycks2016gaussian} activation and LayerNorm~\citep{ba2016layernormalization}. 
We consider two projector architectures: a 2-layer and a 3-layer MLP. Architectural details are summarized in Table~\ref{tab:encoder_results} and Section~\ref{sec:training_subsection}.

\subsection{Cross-Entropy Issue}
\label{sec:entropy}

Prior work in both textual and multimodal compression typically employs \emph{two-stage} training. 
In the first stage, the projection layer is pretrained on an auxiliary alignment task while the large language model (LLM) is frozen; only the projector is updated. In the second stage, either the projector alone or both the projector and LLM are fine-tuned on the downstream task. 
In contrast, CoRoVA adopts a \emph{single-stage training}, omitting the pretraining stage. We tried to pretrain the projector, but found no performance gain (Sec.~\ref{sec:training_subsection}).

Most prior work on the related task — training a projection from encoder outputs into the embedding space of an LLM — has relied on instruction-tuned models and QA datasets, and trained primarily with cross-entropy loss~\citep{pmlr-v139-jaegle21a, liu2023visualinstructiontuning, zemskova20253dgraphllmcombiningsemanticgraphs}. There are, however, notable exceptions. For example, xRAG incorporated KL divergence loss \citep{cheng2024xragextremecontextcompression}, reporting that it had a greater impact on downstream performance than cross-entropy (CE) loss. Another deviation from pure CE training is Flamingo~\citep{alayrac2022flamingovisuallanguagemodel}, which employed the two-term contrastive loss introduced in~\citet{pmlr-v139-radford21a}.

Cross-entropy (negative log-likelihood) is the standard objective for training autoregressive LLMs: it measures how well the model’s predicted next-token distribution matches the target distribution. The formula for cross-entropy loss is the following:
$$
\mathcal{L}_{CE}(\theta)
= -\frac{1}{T}\sum_{t=1}^{T}\log p_{\theta}(y_t|y_1, \dots, y_{t-1}).
$$
Our loss ablation study (Table~\ref{tab:loss_ablation}) reveals that cross-entropy loss alone is \emph{insufficient for optimizing repository-level code completion}, as it exhibits weak correlation with our target EM and ES metrics. 

EM measures the percentage of predictions that exactly match the groundtruth -- a strict criterion that rewards only fully correct completions. In contrast, ES quantifies similarity via the minimum edit distance between prediction and reference~\cite{tan2024promptbasedcodecompletionmultiretrieval, repoformer}, providing a softer signal that rewards partial correctness.

The misalignment arises because CE is inherently a \emph{token-level} objective: it maximizes the likelihood of each ground-truth token conditioned on the prefix, without explicitly modeling \emph{sequence-level} correctness. As a result, CE fails to account for error propagation in long-horizon generation. An early mistake may still lead to a fluent, syntactically valid continuation that incurs low CE loss, while being semantically incorrect at the repository level.

Consequently, models trained only with CE achieve lower training loss yet underperform on target EM and ES metrics (Table~\ref{tab:loss_ablation}), confirming a misalignment between optimization objective and evaluation metric. These results motivated us to explore methods for \emph{directly optimizing sequence-based metrics}, including approaches from reinforcement learning.

\begin{figure*}[th!]
\setlength{\abovecaptionskip}{4pt}
  \centering

  \begin{subfigure}[b]{0.25\textwidth}
    \centering
    \includegraphics[width=\linewidth]{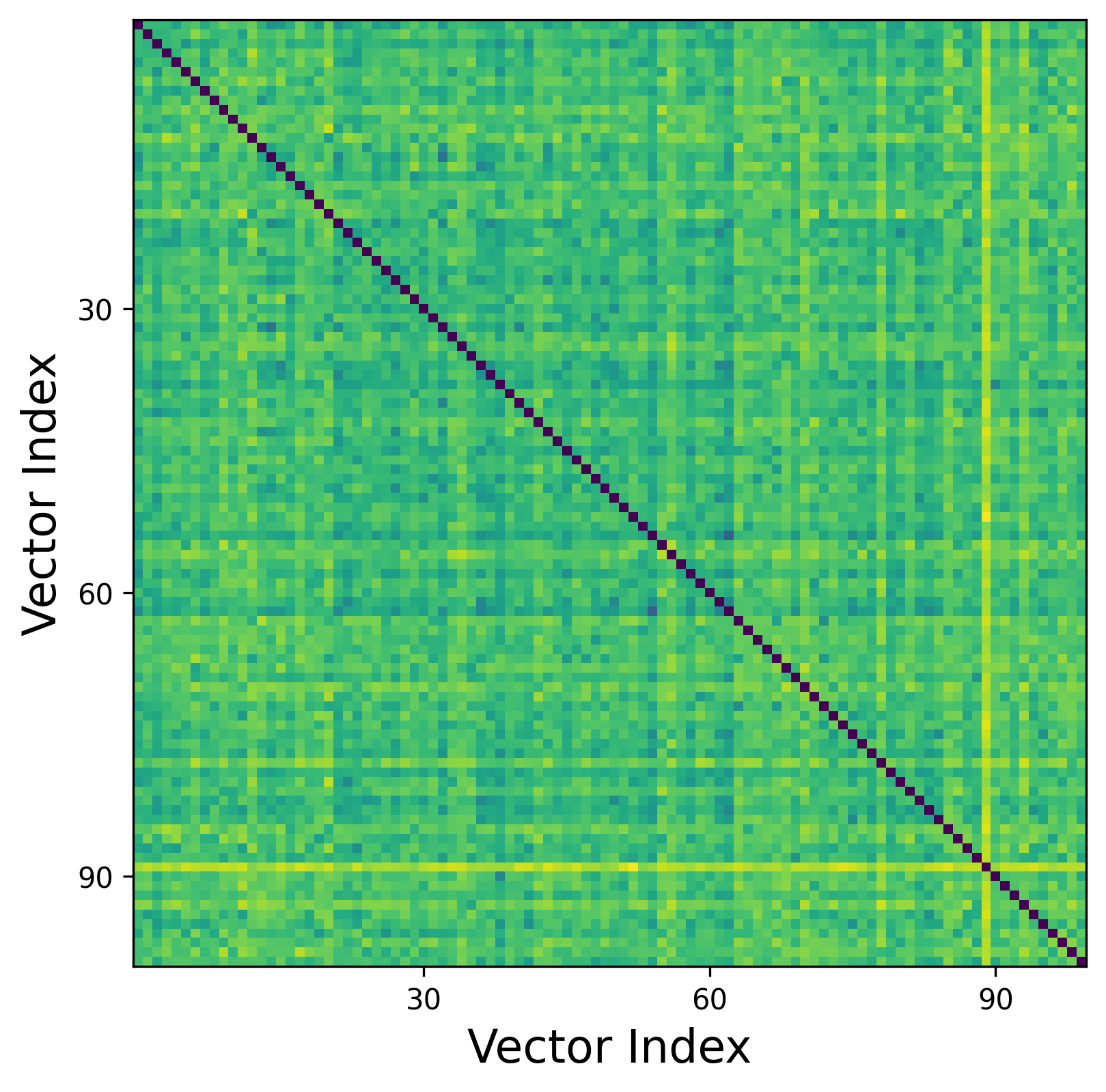}
    \caption{Encoder output}
    \label{fig:cos_qwen}
  \end{subfigure}\hspace{0.05\textwidth}
  \begin{subfigure}[b]{0.25\textwidth}
    \centering
    \includegraphics[width=\linewidth]{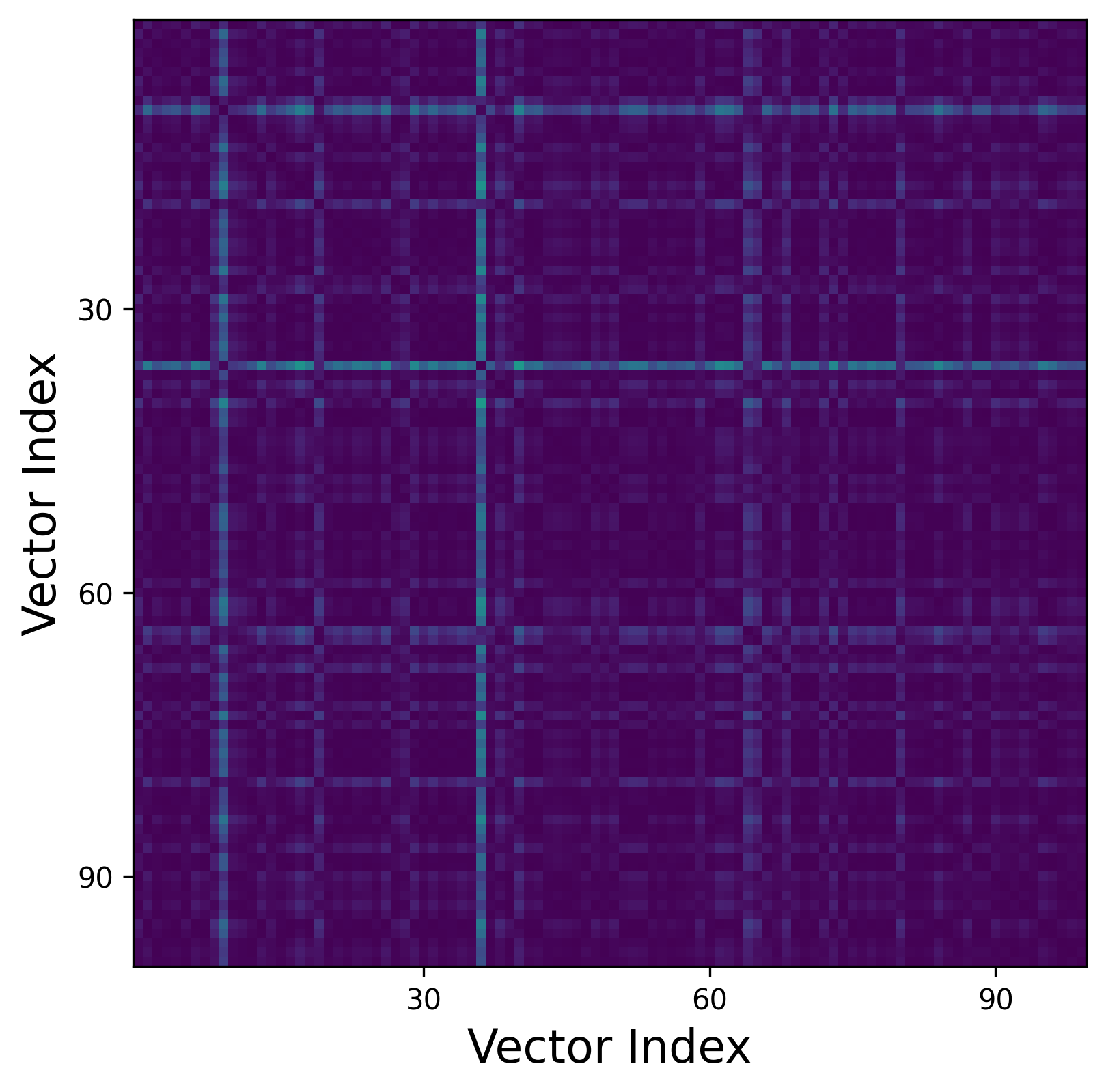}
    \caption{Collapsed projector output}
    \label{fig:cos_bad}
  \end{subfigure}\hspace{0.05\textwidth}
  \begin{subfigure}[b]{0.274\textwidth}
    \centering
    \includegraphics[width=\linewidth]{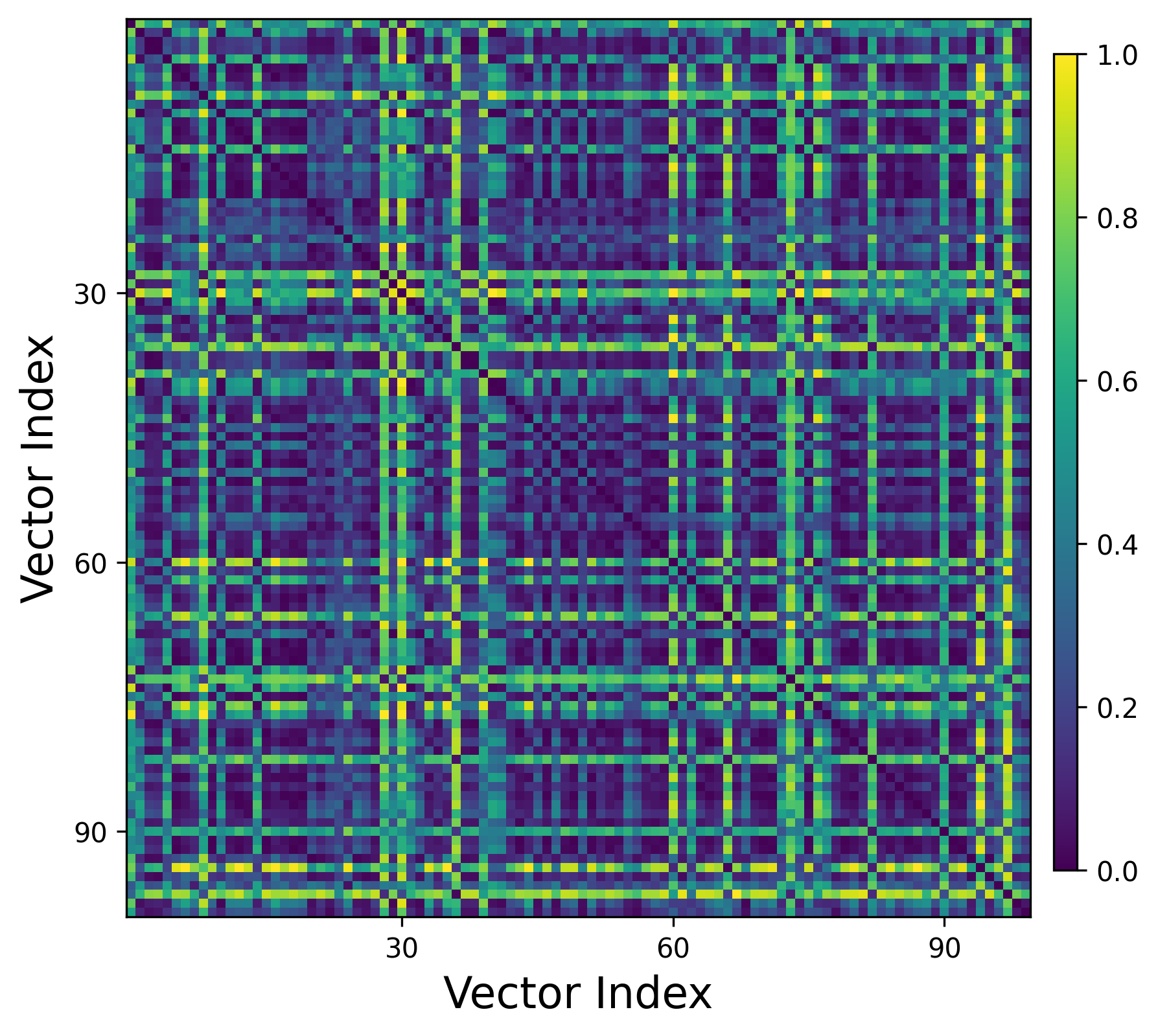}
    \caption{Well-separated projector output}
    \label{fig:cos_good}
  \end{subfigure}

  \caption{Pairwise cosine distances between vector outputs. While the encoder representations remain well-separated (a), the projected vectors may collapse, becoming nearly indistinguishable (b). Introducing the Cosine Alignment loss (Eq.~\ref{eq:cos_loss}) helps preserve the distinctions among the projections, preventing excessive overlap (c).}
  \label{fig:cos}
\end{figure*}

\subsection{RL Loss Component}
\label{sec:scst_loss}

As noted in~\citet{rennie2017selfcriticalsequencetrainingimage}, deep generative models for text are typically trained to maximize the likelihood of the next ground-truth word conditioned on the ground-truth prefix via backpropagation. 
This training paradigm is commonly referred to as teacher forcing~\citep{bengio2015nips}. 
However, it introduces a discrepancy between training and inference: at test time, the model generates each word conditioned on its own previous predictions rather than the ground-truth sequence. 
This exposure bias~\citep{Ranzato2015SequenceLT} can lead to the accumulation of errors during generation, as the model has never been exposed to its own predictions during training.

Our target metrics, Exact Match and Edit Similarity, are inherently affected by teacher-forcing bias, as they evaluate predictions at the sequence level. 
Previous studies have shown that both exposure bias and the non-differentiability of sequence-based evaluation metrics can be mitigated using techniques from reinforcement learning~\citep{rl}. 
In particular,~\citet{Ranzato2015SequenceLT} and~\citet{rennie2017selfcriticalsequencetrainingimage} apply the REINFORCE algorithm~\citep{reinforce} to directly optimize non-differentiable, sequence-level metrics.

Assume we train an LLM decoder model with parameters $\theta$. REINFORCE is based on the observation that the expected gradient of a non-differentiable reward function can be computed as:
\begin{equation}
    \nabla_\theta \mathcal{L}_R(\theta) = - \mathbb{E}_{y \sim p_\theta} \big[ r(y) \, \nabla_\theta \log p_\theta(y) \big],
\end{equation}
where $y = (y_1 , \dots, y_T)$ is a sequence of generated tokens, $y_t \sim p_\theta(y_t | y_{1}, \dots, y_{t-1})$.

In practice, the expected gradient can be approximated using a single Monte-Carlo sample from $p_\theta$. 
Using the sum of our target metrics as a reward function yields:
\begin{equation}
    \mathcal{L}_R(\theta) = - \left( \text{EM}(y) + \text{ES}(y) \right) \, \sum_{t=1}^T \log p_\theta(y_t | y_1, \dots, y_{t-1}),
\end{equation}
where $\text{ES}(y)$ and $\text{EM}(y)$ are the EM and ES metrics computed from a model-generated sequence with a greedy approach.

We adopt Self-Critical Sequence Training (SCST) \cite{rennie2017selfcriticalsequencetrainingimage} to stabilize policy-gradient optimization by using the reward of a greedy rollout as a variance-reducing baseline. Greedy decoding is also the standard inference strategy in code completion, making it a natural reference behavior for learning.
We do not employ PPO-style optimization \cite{schulman2017ppo}, as it requires training an additional value (critic) or reward estimation model, whereas our setting already provides deterministic reward functions. Similarly, GRPO \cite{shao2024deepseekmath} requires computing gradients for multiple sampled rollouts per prompt, which significantly increases computational cost. Given these considerations, SCST offers a practical trade-off between training stability and efficiency while directly optimizing the desired reward signal.




\subsection{Cosine Alignment Loss}
\label{sec:cosine_loss}

During initial experiments on training the projection from encoder representations into the LLM embedding space, we observed that the projection MLP often collapsed to an almost one-dimensional subspace: the angles between projected vectors converged to nearly zero across most pairs (Figure \ref{fig:cos_bad}), even though the encoder itself remained expressive and produced embeddings with pairwise cosine similarities broadly distributed in the range $[0.0,1.0]$ (Figure~\ref{fig:cos_qwen}). 

This behavior is problematic, since our goal is to preserve the distinctions between retrieved text chunks.
To address this collapse and retain the relative differences among encoder embeddings after projection, we introduce a specialized \textit{Cosine Alignment loss}:
\begin{equation}\label{eq:cos_loss}
\mathcal{L}_{A}(\theta) = \frac{1}{B^2} \| S_C(y_{\text{enc}}) - S_C(y_{\text{proj}}) \|_F^2 ,
\end{equation}
where $S_C(\cdot)$ denotes the cosine similarity matrix between vectors of the output batch; $y_\text{enc}$ and $y_\text{proj}$ represent the encoder and projection output batches, respectively; and $B$ is the batch size. This loss enforces preservation of pairwise cosine similarities within a batch by minimizing the mean squared error (MSE) between the similarity matrices. Each matrix has dimensions $B\times B$, so division by $B^2$ computes the mean across elements.


The loss formulation in Eq.~\ref{eq:cos_loss} preserves relative differences between retrieved contexts. 
Figure~\ref{fig:cos_good} shows the resulting cosine-distance matrix for 100 random samples after applying the loss: the projections remain mostly well-separated, and their cosine distance matrix has the same structure as the original encoder embeddings. 
In contrast, without this regularization, the projector outputs collapse into an almost indistinguishable representation (Fig.~\ref{fig:cos_bad}).

\subsection{KL Loss}

In contrast to the recent findings~\cite{cheng2024xragextremecontextcompression}, we observe that additional KL-divergence loss between models trained with compressed versus uncompressed retrieved context does not improve prediction quality. 
This divergence with the results presented in the xRAG~\cite{cheng2024xragextremecontextcompression} paper is notable because xRAG assigns the KL component a relatively large weight (2.0) compared to the negative log-likelihood (1.0), motivated by ablations on instruction-tuned models for QA, where the authors attributed performance gains primarily to the KL-loss component, arguing that it improved the model’s resilience rate -- the fraction of examples that remain correct both before and after retrieval augmentation~\cite{cheng2024xragextremecontextcompression}.


To thoroughly evaluate the xRAG approach to projector training, we perform an ablation study using multiple loss formulations, including a full reproduction of the xRAG loss described above. As shown in Table~\ref{tab:loss_ablation}, this configuration yields performance below the unaugmented baseline, indicating that KL distillation is not a reliable training signal for projector learning in our task.

\begin{figure}
\setlength{\abovecaptionskip}{4pt}
    \centering
    \includegraphics[width=0.98\linewidth]{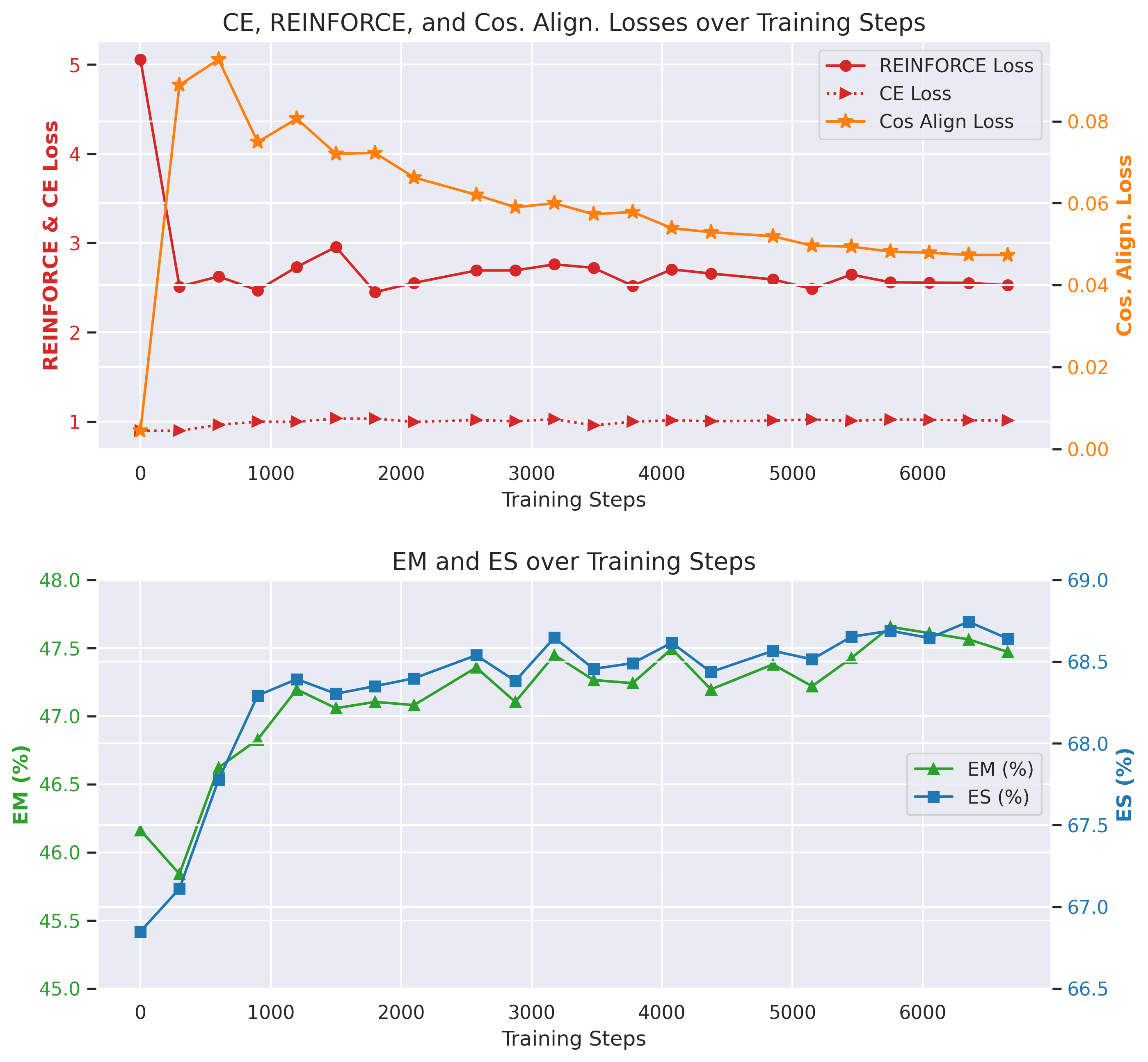}
    \caption{Relationship between the three loss components (Cross-Entropy, REINFORCE, Cosine Alignment) and the evaluation metrics Exact Match (EM) and Edit Similarity (ES).}
    \label{fig:losses_vs_metrics}
\end{figure}

\subsection{Final Loss Function}
\label{sec:loss}

We optimize our model using the following composite loss function: 
\begin{equation}\label{eq:loss}
    \mathcal{L}(\theta) = \alpha_{CE}\mathcal{L}_{CE}(\theta) + \alpha_R \mathcal{L}_R (\theta) + \alpha_A \mathcal{L}_A(\theta),
\end{equation}
where the coefficients $\alpha_{CE}$, $\alpha_{R}$, and $\alpha_{A}$ are weighting factors. These weights are selected through hyperparameter tuning using  Optuna\footnote{\href{https://optuna.org}{https://optuna.org}}~\citep{akiba2019optunanextgenerationhyperparameteroptimization}. 
These and other training hyperparameters for all trained models are listed in Section~\ref{sec:training_subsection}.
The loss dynamic and its correspondence to target metrics EM and ES can be seen in Figure \ref{fig:losses_vs_metrics}.


\section{Experiments}
\label{sec:expers}
This section describes our experimental setup, covering datasets, models, retrieval and compression strategies, evaluation protocols.

\subsection{Datasets and Splitting}
\label{sec:dataset}

For initial ablation studies and experiments with 1.5B-parameter models, we use the Python subset of The Stack~\citep{kocetkov2022stack3tbpermissively}. Following prior work~\cite{repoformer}, we organize files at the repository level and apply quality filters to ensure meaningful cross-file context. Specifically, we remove repositories with fewer than 50 GitHub stars, fewer than five files, or files containing fewer than three import statements. After filtering, this dataset yields 150k code completion examples.

For all main results with the 7B-parameter models, we adopt The Stack v2~\cite{lozhkov2024starcoder2stackv2}, which contains updated Python and Java repositories collected prior to September 14, 2023. To ensure sufficiently rich repository context, we apply stricter project-level constraints: each repository must contain at least ten files, and each file must have a minimum of forty lines of code. We further restrict all data to the main branch of each repository to maintain consistency. To prevent data leakage, repositories are split at the repository level into training, validation, and test sets using a 90/5/5 ratio. For efficient evaluation, we uniformly subsample 4,000 validation examples for inference-time analysis and 4,000 test examples for final metric reporting on the Java subset. Comprehensive dataset statistics for both languages are provided in Table~\ref{tab:dataset_stats}.

    

\begin{table}[t]
\setlength{\abovecaptionskip}{6pt}
\centering
\caption{Statistics of the filtered The Stack v2 dataset. Repositories are split repository-wise to prevent data leakage. We use a subset of 4,000 validation examples for fast inference.}
\label{tab:dataset_stats}

\resizebox{\columnwidth}{!}{%
\begin{tabular}{lrrrr}
    \toprule
    \textbf{Language} &
    \textbf{Total Repositories} &
    \textbf{Train Examples} &
    \textbf{Val. Examples} &
    \textbf{Test Examples} \\
    \midrule
    Python & 118,562 & 2,839,479 & 157,969 & 160,836 \\
    Java   & 226,447 & 5,273,006 & 293,694 & 294,715 \\
    \bottomrule
\end{tabular}%
}
\end{table}

\subsection{Experimental Setup}

\subsubsection{Task Formulation}
Code completion is formulated as a fill-in-the-middle (FIM) task. Given the left context (lines preceding the target code chunk) and the right context (lines following it) from a source file, the model must generate the missing middle segment.
Each target segment consists of $n_t$ consecutive lines ($1 \leq n_t \leq 9$), with $n_t$ sampled from a Poisson distribution to encourage diversity in completion length. This setup reflects realistic code editing scenarios while avoiding bias toward fixed-length completions. 

\subsubsection{Cross-File Context Construction.}

To incorporate repository-level information, we retrieve relevant context from non-target files within the same repository. All such files are segmented into overlapping chunks of $10 \times n_t$ lines, with an overlap of $5 \times n_t$ lines. From this pool, we select the top 10 most relevant chunks based on Jaccard similarity over tokenized code.

We evaluated several retrieval strategies for selecting cross-file context, including sparse methods (BM25 and Jaccard) and dense retrievers based on cosine similarity between embeddings from UniXCoder~\cite{guo2022unixcoderunifiedcrossmodalpretraining} and Jina v2~\cite{gunther2023jina2}. Each retriever-augmented model was compared against a baseline without cross-file context on a subset of 1,600 completion tasks from The Stack. Results are reported in Table~\ref{tab:rag_strats}. Overall, Jaccard similarity and UniXCoder embeddings achieve the strongest performance. Given its simplicity and lower inference latency, we adopt Jaccard similarity as our default retrieval method throughout all experiments.

\begin{table}[]
\setlength{\abovecaptionskip}{5pt}
\caption{Comparison of different retrieval strategies}
\centering
\resizebox{\columnwidth}{!}{%
\begin{tabular}{lccccc}
\hline
\textbf{Metric} & \textbf{W/o CFC} & \textbf{BM25} & \textbf{Jaccard} & \textbf{UniXCoder~\cite{guo2022unixcoderunifiedcrossmodalpretraining}} & \textbf{Jina v2~\cite{gunther2023jina2}} \\
\hline
EM & 50.50 & 55.56 & \textbf{56.19} & \underline{56.00} & 54.31 \\
ES & 73.12 & 76.5 & \underline{76.68} & \textbf{76.84} & 75.56 \\
\hline
\end{tabular}
}
\label{tab:rag_strats}
\end{table}


\subsubsection{Evaluation Protocol}

During training, we use the full available target length, enabling the model to learn multi-line generation.

For evaluation on the validation subset, we focus on single-line completion. This setting offers a more controlled and reproducible assessment of sequence-level correctness, as EM and ES metrics become increasingly ambiguous for longer, multi-line targets. Also, restricting evaluation to single-line completions reduces inference cost without compromising training generality.

Before computing EM and ES metrics, we strip leading and trailing whitespace from predictions and targets to avoid \emph{artificial metric inflation} from trivially predictable formatting tokens (e.g., indentation). This normalization ensures that EM and ES primarily reflect code-level correctness rather than superficial whitespace accuracy.


\begin{table}[t!]
\setlength{\abovecaptionskip}{6pt}
\caption{Ablation study of projection-training approaches. Results without context augmentation are denoted \textnormal{w/o CFC}; results with uncompressed cross-file context \textnormal{(w/ CFC)} are highlighted in gold. $\alpha_{KL}$ denotes the weight of the KL loss, other loss terms follow Section~\ref{sec:loss}. Metrics are reported on an evaluation subset prepared from The Stack ($4.3k$ samples).}
\centering
\resizebox{\columnwidth}{!}{%
\begin{tabular}{l c c c c c c c}
\hline
\textbf{Method} & \textbf{$\alpha_{CE}$} &  \textbf{$\alpha_{R}$} & \textbf{$\alpha_{A}$} & \textbf{$\alpha_{KL}$} & \textbf{CE Loss $\downarrow$} & \textbf{EM $\uparrow$} & \textbf{ES $\uparrow$} \\
\hline
Base Model w/o CFC & - & - & - & - & 0.97 & 45.97 & 66.57 \\
\rowcolor{mygold}
Base Model w/ CFC & - & - & - & - & 0.99 & 50.87 & 69.43 \\
\hline
LLaVA~\citep{liu2023visualinstructiontuning} & 1.0 & 0.0 & 0.0 & 0.0 & \textbf{0.80} & 38.57 & 63.6 \\
REINFORCE-only & 0.0 & 1.0 & 0.0 & 0.0 & 5.18 & 40.61 & 63.91 \\
CE + Cos. Align. & 0.9 & 0.0 & 0.1 & 0.0 & 0.89 & 42.3 & 64.43 \\
xRAG~\citep{cheng2024xragextremecontextcompression} & 1.0 & 0.0 & 0.0 & 2.0 & \underline{0.84} & \underline{44.0} & \underline{65.5}\\
\textbf{CoRoVA (ours)} & 0.9 & 0.1 & 0.1 & 0.0 & 1.02 & \textbf{47.66} & \textbf{68.74} \\
\hline
 \end{tabular}
 }
\label{tab:loss_ablation}
\end{table}

\subsection{Training}
\label{sec:training_subsection}

\subsubsection{Training Details.}

We train 2- and 3-layer MLP projection modules that map sentence encoder outputs (e.g., UniXCoder or Qwen3-Embedding) into the dimension of code LLM embeddings.
For each sample, the top-10 cross-file contexts, encoded and projected into LLM representations at the time of RAG database creation, are concatenated with the code completion prompt embeddings before being passed to the LLM (Fig.~\ref{fig:all}b). 
When comparing against the LLM with non-compressed text context, the top-10 retrieved contexts are concatenated into one sequence, truncated to 512 tokens, and then concatenated with the same code completion prompt (Fig.~\ref{fig:all}a). The code completion prompt budget (without retrieved context) is 2,000 tokens for both methods. 
As a result, the input sequence length in CoRoVA is 502 tokens shorter than in conventional RAG. 

During training, only the projection weights are updated, while both the encoder and the LLM remain frozen. 
Optimization is performed using the joint loss described in Section \ref{sec:loss}, which combines all three loss components.
Cross-entropy is only computed over the sequence after the \texttt{<|fim\_middle|>} special token.
For REINFORCE loss, we generate 50 tokens using greedy decoding and evaluate EM and ES metrics on the obtained sequence. 

For our primary evaluations, we used the Qwen2.5-Coder family of models, with the Qwen3-Embedding-0.6B model serving as the encoder. 
A three-layer MLP was employed as the projector, mapping from the encoder dimension to twice the embedding size of the LLM, and finally down to the LLM’s embedding size.
A GELU activation and a LayerNorm were applied between the first and second layers, and again between the second and final layer.
Training hyperparameters for different model sizes are described in Table~\ref{tab:training_parameters}.

\begin{table}[t!]
\setlength{\abovecaptionskip}{5pt}
\caption{Optimal hyperparameter set for projection training}
\centering
\resizebox{0.75\columnwidth}{!}{%
\begin{tabular}{l c c }
\hline
\textbf{Hyperparameter} & \textbf{1.5B models} & \textbf{7B models} \\
\hline
Optimizer & AdamW & AdamW \\
$\alpha_{A}$ (Cosine Alignment) & 0.1 & 0.2 \\
$\alpha_{CE}$ (Cross-Entropy) & 0.9 & 0.9\\
$\alpha_{R}$ (REINFORCE) & 0.1 & 0.05\\
Learning rate (LR) & 1e-3 & 5e-5\\
LR scheduler type & cosine & cosine\\
Warmup ratio & 0.03 & 0.04 \\
Weight decay & 0.0 & 0.0 \\
\#Epochs & 3 & 0.12\\
Effective batch size & 66 & 64 \\
\#Train samples & 150k  & 2.8M\\
\hline
\end{tabular}
}
\label{tab:training_parameters}
\end{table}

\begin{table}[t!]
\setlength{\abovecaptionskip}{6pt}
\caption{Results on code-completion benchmarks. All methods use Qwen2.5-Coder-7B as the base code-generating LLM. For Java on The Stack v2, we use the test split of the dataset.}
\centering
\resizebox{\columnwidth}{!}{%
\begin{tabular}{c c c c c c c}
\toprule
\textbf{Language} & \textbf{Benchmark} & \textbf{Model} & \textbf{Seq. Len.} & \textbf{EM $\uparrow$} & \textbf{ES $\uparrow$} & \textbf{CodeBLEU$\uparrow$}\\
\midrule
\multirow{12}{*}{Python} & \multirow{3}{*}{CCEvalLong~\cite{repoformer}} & w/o CFC & 2000 & 45.64 & 71.75 & 55.82 \\
& & w/ CFC & 2512 & \textbf{49.74} & \underline{73.2} & \textbf{58.01} \\
& & CoRoVA (ours) & 2010 & \underline{47.16} & \textbf{73.35} & \underline{57.46} \\
\cline{2-7}
& \multirow{6}{*}{RepoEval~\cite{repocoder}} & w/o CFC & 2000 & 58.13 & 77.98  & 58.09\\
& & w/ CFC & 2512 & \textbf{64.56} & \textbf{81.69} & \textbf{64.34} \\
& & CoRoVA (ours) & 2010 & \underline{60.56} & \underline{80.46} & \underline{59.67}\\
& & Context Pruning & 2010 & 57.91 & 78.1 & 58.05 \\
& & Context Summar. & 2010 & 58.04 & 78.26 & 58.23 \\
& & CodePromptZip~\cite{he2025codepromptzipcodespecificpromptcompression} & 2010 & 57.91 & 78.3 & 58.26 \\
\cline{2-7}
& \multirow{3}{*}{RepoEval Api~\cite{repocoder}} & w/o CFC & 2000 & 48.75 & 74.13 & 52.32 \\
& & w/ CFC & 2512 & \textbf{55.00} & \textbf{79.31} & \textbf{57.60} \\
& & CoRoVA (ours) & 2010 & \underline{51.31} & \underline{77.95} & \underline{55.56} \\
\cline{1-7}
\multirow{6}{*}{Java} & \multirow{3}{*}{The Stack v2~\cite{lozhkov2024starcoder2stackv2}} & w/o CFC & 2000 & 68.3 & 82.8 & 63.96 \\
 &  & w/ CFC & 2512 & \underline{70.92} & \underline{84.43} & \underline{65.69} \\
 &  & CoRoVA (ours) & 2010 & \textbf{71.2} & \textbf{84.87} & \textbf{66.03} \\
\cline{2-7}
 & \multirow{3}{*}{RepoEval-Updated~\cite{liu2024graphcoderenhancingrepositorylevelcode}} & w/o CFC & 2000 & 52.71 & 76.49 & 64.13 \\
 &  & w/ CFC & 2512 & \textbf{55.63} & \textbf{78.03} & \textbf{66.01} \\
 &  & CoRoVA (ours) & 2010 & \underline{54.11} & \underline{77.21} & \underline{65.27} \\
\bottomrule
 \end{tabular}
 }
\label{tab:res_bench}
\end{table}


\subsubsection{Single-stage vs. Two-stage Training.}
Whereas most prior work adopts two-stage training, we use a single-stage pipeline based on a composite loss function, discussed in Section~\ref{sec:loss}. For completeness, we also evaluated a conventional two-stage pretrain-finetune pipeline for projection training.

In prior work, pretraining often relies on parallel datasets, such as paraphrase pairs in xRAG or image-caption pairs in LLaVA. Inspired by xRAG, we experimented with a similar pretraining approach, attempting to reconstruct retrieved context chunks from projected vectors by optimizing cross-entropy loss. This approach did not yield improvements in the second stage of training, likely due to the entropy issues discussed in Section~\ref{sec:entropy}.

\subsubsection{Pretraining of the projector.}
The work~\cite{kuratov2025cramming1568tokenssingle} demonstrates that up to 1,568 tokens can be compressed into a single continuous ``memory'' token by treating the token as a trainable parameter and optimizing it via backpropagation with a cross-entropy reconstruction loss. Because these continuous tokens reconstruct to reference texts, we treat them as ground truth for training our projection layer. We encode text with our encoder, project the resulting embeddings into a single token, and optimize a mixture of mean squared error (MSE) and cosine-similarity (CS) losses between the projected embedding and the trained ground-truth compressed token.

However, the space spanned by the memory tokens proved to be highly non-smooth. For instance, identical text inputs could be compressed into vectors that are widely separated, and introducing even small perturbations to a learned memory token often results in the reconstruction of completely different text. This leads to poor generalization for an MLP module attempting to map into this space. Consequently, learning a projection into such a space requires extreme overparameterization, effectively amounting to memorizing the entire dataset. As a result, we could only overfit on a small subset of memory tokens and were unable to learn a meaningful translation into the memory token space.
We leave the more sophisticated pretraining of the projection module for code compression to future work.


\subsection{Main Results}

\subsubsection{Overall Benchmark Performance}

Table~\ref{tab:res_bench} presents our main results across multiple established code completion benchmarks. We evaluate CoRoVA, our approach for retrieved-context compression, against two key baselines that isolate the effect of cross-file context usage: (i) \textbf{w/o CFC} -- uses only local in-file context (no cross-file retrieval); (ii) \textbf{w/ CFC} -- incorporates the full set of retrieved cross-file context (CFC) without compression (Figure~\ref{fig:all}a).

All models share the same underlying LLM (Qwen2.5-Coder-7B) and retrieval mechanism, they differ solely in how retrieved context is integrated, enabling a clean ablation of compression efficiency.

Despite the negligible latency impact introduced by the additional 10 tokens, our approach surpasses the w/o CFC baseline on EM and ES metrics by a sizable margin, which makes CoRoVA preferable in latency-limited environments, such as IDE code completion, where vanilla RAG introduces noticeable latency impact in the range of 20-38\%.
Detailed latency measurements are presented in Section~\ref{sec:practical} and in Tables~\ref{tab:tpot_transformers_vllm_combined}, \ref{tab:dissagr_cmp_prefill_decode_full}. 


\begin{figure}
\setlength{\abovecaptionskip}{2pt}
    \centering
    \includegraphics[width=0.71\linewidth]{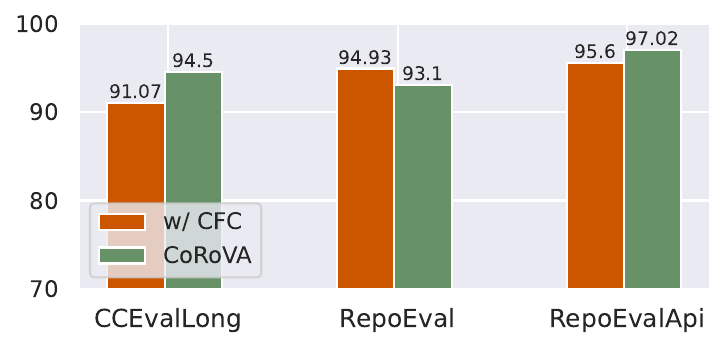}
    \caption{Resilience rate of two augmentation methods -- uncompressed retrieved cross-file context (w/ CFC) and CoRoVA -- relative to a Qwen2.5-Coder-7B baseline without RAG.}
    \label{fig:resilience}
\end{figure}

\begin{table}[t!]
\setlength{\abovecaptionskip}{5pt}
\caption[Performance of xRAG on RepoEval benchmark]{Performance of xRAG\footnotemark{} on RepoEval benchmark}
\centering
\resizebox{0.85\columnwidth}{!}{%
\begin{tabular}{l c c c c}
\hline
\textbf{Model} & \textbf{Sequence Length}  & \textbf{EM $\uparrow$} & \textbf{ES $\uparrow$} & \textbf{CodeBLEU$\uparrow$} \\
\hline
w/ CFC & 2000 & 0.19 & 14.05 & 3.88 \\
xRAG & 2010 & 0.19 & 13.42 & 4.10 \\
\hline
\end{tabular}
}
\label{tab:xrag_results}
\end{table}
\footnotetext{\href{https://huggingface.co/Hannibal046/xrag-7b}{https://huggingface.co/Hannibal046/xrag-7b}}

Table~\ref{tab:examples} presents examples illustrating how CoRoVA handles code completion tasks from the benchmarks, compared with both the baseline model without cross-file context and the model using uncompressed cross-file context.

\subsubsection{Comparison to Compression Baselines}

On RepoEval, we compare CoRoVA against token-space compression approaches that target a comparable context budget, including token pruning -- compression of the context into a fixed token budget by selecting the most representative tokens; and summarization -- model is prompted to produce a maximally faithful fixed-length summary of the retrieved cross-file context (Table~\ref{tab:res_bench}).
Both pruning and summarization are performed using Qwen2.5-Coder-7B Instruct.
We additionally include CodePromptZip~\citep{he2025codepromptzipcodespecificpromptcompression}, a code-specific prompt compression method.
Across these baselines, we observe that methods operating in token space \emph{fail to preserve completion quality at the extreme compression level} required here, whereas CoRoVA \emph{retains the benefits of cross-file context}.

Table~\ref{tab:xrag_results} reports results for the original xRAG pipeline on RepoEval.
We do not include xRAG in the main comparison in Table~\ref{tab:res_bench} because it relies on a general-purpose Mistral-7B model and a compression mechanism trained on retrieved natural-language documents for downstream QA.
Nevertheless, the results suggest that document-oriented compression objectives do not translate well to code completion, yielding at best marginal gains in CodeBLEU~\cite{ren2020codebleu} while degrading other metrics such as Edit Similarity~\cite{repocoder}.

Overall, under \emph{extreme context budgets}, CoRoVA consistently \emph{provides stronger gains} than existing token-space and document-style compression baselines for code completion.

\subsubsection{Robustness to Retrieval Augmentation}

Figure~\ref{fig:resilience} shows the resilience rate -- the percentage of instances where the model’s response remains correct both before and after retrieval augmentation~\cite{cheng2024xragextremecontextcompression} -- for CoRoVA compared to a model using uncompressed cross-file context. Overall, the rates are comparable, with CoRoVA showing a slight advantage on average.

\subsubsection{Execution-based Evaluation}

To rigorously evaluate the functional correctness of generated code beyond syntactic validity, we adopt the execution-based Pass@k metric~\cite{bench_humaneval_chen2021evaluatinglargelanguagemodels}. We select the RepoEval Function Completion benchmark~\cite{repocoder}, because each function has \emph{manually verified test suites} with strict coverage requirements, unlike RepoEval’s line/API completion subsets, where tests frequently lack coverage guarantees for completion targets.

While open-source Pass@k evaluation frameworks exist\footnote{\url{https://github.com/Elendil3703/RepoEval_Exec}}, we encountered irreproducible test environments and dependency conflicts when attempting to reuse them. To ensure evaluation integrity, we implemented our own pipeline to test code completion metrics.

From the benchmark’s 8 repositories, only \texttt{leopard/ai-betty}, \texttt{deepmind/tracr}, and \texttt{amazon-science/patchcore-inspection} passed all unit tests using their original, unmodified code. We exclude the other five repositories because their test suites are broken, which could skew correctness measurements, resulting in a benchmark of 244 functions. Table~\ref{tab:res_pass} reports Pass@5 results. CoRoVA achieves state-of-the-art results on \texttt{patchcore-inspection} (+3.1\% over best baseline) and \texttt{ai-betty} (+1.6\%), demonstrating effective context utilization. For \texttt{tracr}, all context-aware methods underperform the no-context baseline (w/o CFC), suggesting repository-specific sensitivity to cross-file dependencies, but CoRoVA still outperforms the w/ CFC baseline.

\begin{table}[t!]
\setlength{\abovecaptionskip}{6pt}
\caption{Pass@5 (\%) on RepoEval Function Completion benchmark, reported per repository. Qwen2.5-Coder-7B is used as the base code-generating LLM for all methods. Repositories: \texttt{amazon-science/patchcore-inspection}, \texttt{deepmind/tracr}, \texttt{leopard/ai-betty}.}
\centering
\resizebox{\columnwidth}{!}{%
\begin{tabular}{l l  c  c  c }
\hline 
\textbf{Model} & \textbf{Seq. Length} & \textbf{patchcore-inspection} & \textbf{tracr} & \textbf{ai-betty} \\
\hline
w/o CFC & 2000 & \underline{64.92} & \textbf{56.11} & 63.14\\
w/ CFC & 2512 & 63.09 & 52.39 & \underline{63.97}\\
CoRoVA (ours) & 2010 & \textbf{66.94} & \underline{52.76} & \textbf{64.99}\\
\hline
 \end{tabular}
 }
\label{tab:res_pass}
\end{table}

\subsubsection{Objective Ablation}

Table~\ref{tab:loss_ablation} presents the results of ablation studies across different loss formulations, comparing four configurations: Cross-Entropy only (LLaVA-style), REINFORCE only, Cross-Entropy with Cosine Alignment, and Cross-Entropy with KL Loss (the xRAG objective).
As discussed in Section~\ref{sec:entropy}, relying solely on the Cross-Entropy objective degrades performance on both EM and ES metrics.
Conversely, optimizing exclusively with the REINFORCE loss leads to uncontrolled entropy growth and fails to outperform the w/o CFC baseline, due to the absence of a variance-reducing baseline. 
In contrast, only a carefully balanced combination of all three loss components (Section~\ref{sec:loss}) yields consistent improvements across the target metrics (Figure~\ref{fig:losses_vs_metrics}).

\subsubsection{Architecture Ablation}
\label{sec:encoder_ablation}

The ablation in Table~\ref{tab:encoder_results} studies the effect of encoder choice and projection depth. We evaluate two encoders and code modalities: UniXCoder~\citep{guo2022unixcoderunifiedcrossmodalpretraining} with AST representations of retrieved code, and the Qwen3-Embedding-0.6B~\cite{zhang2025qwen3embeddingadvancingtext} model with retrieved code. Qwen3-Embedding-0.6B, used as the retrieved-context compressor, outperforms UniXCoder. A three-layer MLP projection further improves both EM and ES but increases the number of trainable parameters by roughly $4\times$.

\begin{table}[h!]
\setlength{\abovecaptionskip}{5pt}
\caption{Examples of generated code for the code-completion benchmarks from Table~\ref{tab:res_bench}. Although some benchmark tasks involve multi-line completions, only one-line cases are displayed to fit more examples.}
\centering
\resizebox{\columnwidth}{!}{%
\begin{tabular}{l c}
\hline
\textbf{Method} & \textbf{Code Completion}\\
\hline
ground truth & \texttt{pos += slice\_length} \\
w/o CFC & \texttt{return} \\
w/ CFC & \texttt{yield ""} \\
CoRoVA & \texttt{pos += slice\_length} \\
\hline
ground truth & \texttt{fields\_values[name] = NAO} \\
w/o CFC & \texttt{if name in fields:} \\
w/ CFC & \texttt{if name in fields:} \\
CoRoVA & \texttt{fields\_values[name] = NAO} \\
\hline
ground truth & \texttt{trial\_metadata: Iterable[UnitMetadataUpdate],} \\
w/o CFC & \texttt{trial\_metadata: Iterable[} \\
w/ CFC & \texttt{trial\_metadata: Iterable[UnitMetadataUpdate],} \\
CoRoVA & \texttt{trial\_metadata: Iterable[key\_value\_pb2.KeyValue],} \\
\hline
ground truth & \texttt{from jax import random} \\
w/o CFC & \texttt{import random} \\
w/ CFC & \texttt{from jax import random} \\
CoRoVA & \texttt{from jax import random} \\
\hline
ground truth & \texttt{from fortuna.prob\_model.posterior.map.map\_stat} \\
w/o CFC & \texttt{cfg = compile\_config(cfg, create\_cfg=create\_cfg)} \\
w/ CFC & \texttt{from fortuna.prob\_model.posterior.map.map\_ste} \\
CoRoVA & \texttt{\# Get env\_fn from env\_setting.} \\
\hline
\end{tabular}
}
\label{tab:examples}
\end{table}

\begin{table}[h!]
\setlength{\abovecaptionskip}{6pt}
\caption{Comparison of encoder and projector configurations, including trainable parameter counts and downstream performance. Parameter differences arise from the encoder embedding dimensionality and the MLP depth. All configurations are trained for approximately 6,600 steps (3 epochs).}
\centering
\resizebox{\columnwidth}{!}{%
\begin{tabular}{l l l c c c}
\hline
\textbf{Encoder} & \textbf{Modality} & \textbf{Projection} & \textbf{\#Trainable Params.} & \textbf{EM} & \textbf{ES} \\
\hline
UniXCoder & AST & 2-layer MLP & \hphantom{0}3.5M & 46.69 & 67.65 \\
UniXCoder & AST & 3-layer MLP & 16.5M & 46.94 & \underline{68.26} \\
Qwen3-Embedding & Code & 2-layer MLP & \hphantom{0}3.9M & \underline{47.01} & 68.15 \\
Qwen3-Embedding & Code & 3-layer MLP & 17.3M & \textbf{47.66} & \textbf{68.74} \\
\hline
\end{tabular}
}
\label{tab:encoder_results}
\end{table}

\subsection{Online Inference Latency (TTFT/TPOT)}
\label{sec:practical}

CoRoVA pipeline shifts the costly \emph{context understanding} computation off the critical inference path.
During repository indexing (i.e., when the IDE builds or refreshes the retrieval database), we run the Encoder + Projector on code chunks and store the resulting compressed representations.
At inference time, the system only appends these precomputed representations to the prompt, so the primary factor affecting user-perceived latency is the \emph{effective sequence length}.
We quantify the one-time indexing overhead separately in Section \ref{sec:rag_build_cost}, and in this section, we focus on the online latency benefits of sequence length reduction provided by CoRoVA.

Two deployment patterns dominate today’s LLM serving landscape. First, \textit{prefill-decode mixing} uses a single engine that interleaves chunks from the prompt prefill with decoding passes across requests. For instance, one of the engines, which utilizes this approach, is vLLM framework~\cite{kwon2023pagedattention}.
Second, \textit{disaggregated prefill-decode}, when prefill and decode run on separate GPU pools or nodes (possibly on different clusters) with independent resource plans. An example of an engine that uses this approach is DistServe~\cite{zhong2024distserve}.

Colocating prefill and decode is utilization-friendly and achieves high throughput on single machines via memory-efficient KV management and continuous batching.
However, prefill and decode contend for distinct resources and interfere with each other, which makes it hard to independently control TTFT (\emph{time-to-first-token}) and TPOT (\emph{time-per-output-token}) under enterprise's Service Level Agreement (SLA). As a result, systems are often over-provisioned with hardware to satisfy both metrics~\cite{agrawal2024sarathiserve,wang2024revisiting}.

Separating the phases decouples resource allocation and parallelism strategies, eliminating prefill–decode interference and enabling direct tuning of TTFT (prefill stage) and TPOT (decode stage). Operationally, it simplifies capacity planning and horizontal scaling because each fleet can scale along its own bottleneck.
User will operate over IDE in interactive manner, so TTFT of CodeLLM is the main metric to which the experience is sensitive, since, as soon as tokens start generating, user can start reviewing code suggestions. 

For disaggregated serving (transformers) and colocated prefill-decode (vLLM), the results are shown in Table~\ref{tab:dissagr_cmp_prefill_decode_full}. For performance measurements, we report scaling metrics for both inference patterns. 
For benchmarking, we implement separate prefill and decode workers using the transformers runtime \cite{wolf2020transformers}. TPOT measurements are shown in Table~\ref{tab:tpot_transformers_vllm_combined}, and latency reduction measurements for the prefill-only regime (1-token generation) are reported in Table~\ref{tab:prefill_combined}.

\begin{table}
\setlength{\abovecaptionskip}{5pt}
\caption{Latency reduction in prefill-only regime (generation of 1 token) for \textnormal{transformers} and \textnormal{vLLM} frameworks}
\centering
\resizebox{\columnwidth}{!}{%
\begin{tabular}{llcc}
\hline
\textbf{Seq. compression} & \textbf{Model} & \textbf{TTFT (Transformers)} & \textbf{TTFT (vLLM)} \\
\hline
2512$\rightarrow$2010 \textcolor{darkgreen}{$\downarrow$ 20\%} & Qwen2.5-Coder-1.5B &
198.2 $\rightarrow$ 159.3 \textcolor{darkgreen}{$\downarrow$ 20\%} &
\hphantom{1}79.1 $\rightarrow$ \hphantom{1}67.0 \textcolor{darkgreen}{$\downarrow$ 15\%} \\
2512$\rightarrow$2010 \textcolor{darkgreen}{$\downarrow$ 20\%} & Qwen2.5-Coder-7B &
668.1 $\rightarrow$ 539.1 \textcolor{darkgreen}{$\downarrow$ 19\%} &
197.4 $\rightarrow$ 165.9 \textcolor{darkgreen}{$\downarrow$ 16\%} \\
2512$\rightarrow$2010 \textcolor{darkgreen}{$\downarrow$ 20\%} & Qwen2.5-Coder-14B &
820.9 $\rightarrow$ 661.1 \textcolor{darkgreen}{$\downarrow$ 19\%} &
351.5 $\rightarrow$ 291.2 \textcolor{darkgreen}{$\downarrow$ 17\%} \\
\hline
2000$\rightarrow$1510 \textcolor{darkgreen}{$\downarrow$ 24\%} & Qwen2.5-Coder-1.5B &
159.9 $\rightarrow$ 121.2 \textcolor{darkgreen}{$\downarrow$ 24\%} &
\hphantom{0}65.9 $\rightarrow$ \hphantom{0}56.4 \textcolor{darkgreen}{$\downarrow$ 14\%} \\
2000$\rightarrow$1510 \textcolor{darkgreen}{$\downarrow$ 24\%} & Qwen2.5-Coder-7B &
539.1 $\rightarrow$ 406.3 \textcolor{darkgreen}{$\downarrow$ 25\%} &
164.4 $\rightarrow$ 135.8 \textcolor{darkgreen}{$\downarrow$ 17\%} \\
2000$\rightarrow$1510 \textcolor{darkgreen}{$\downarrow$ 24\%} & Qwen2.5-Coder-14B &
660.6 $\rightarrow$ 495.8 \textcolor{darkgreen}{$\downarrow$ 25\%} &
290.0 $\rightarrow$ 240.9 \textcolor{darkgreen}{$\downarrow$ 17\%} \\
\hline
1500$\rightarrow$1010 \textcolor{darkgreen}{$\downarrow$ 33\%} & Qwen2.5-Coder-1.5B &
120.7 $\rightarrow$ \hphantom{0}75.5 \textcolor{darkgreen}{$\downarrow$ 37\%} &
\hphantom{0}56.4 $\rightarrow$ \hphantom{0}48.7 \textcolor{darkgreen}{$\downarrow$ 14\%} \\
1500$\rightarrow$1010 \textcolor{darkgreen}{$\downarrow$ 33\%} & Qwen2.5-Coder-7B &
405.4 $\rightarrow$ 281.0 \textcolor{darkgreen}{$\downarrow$ 31\%} &
136.5 $\rightarrow$ 104.6 \textcolor{darkgreen}{$\downarrow$ 23\%} \\
1500$\rightarrow$1010 \textcolor{darkgreen}{$\downarrow$ 33\%} & Qwen2.5-Coder-14B &
494.9 $\rightarrow$ 339.6 \textcolor{darkgreen}{$\downarrow$ 31\%} &
240.2 $\rightarrow$ 191.4 \textcolor{darkgreen}{$\downarrow$ 20\%} \\
\hline
\end{tabular}
}
\label{tab:prefill_combined}
\end{table}

\begin{table}
\setlength{\abovecaptionskip}{5pt}
\caption{Impact of context compression on TPOT on an NVIDIA A100 for \textnormal{transformers} (disaggregated prefill-decode) and \textnormal{vLLM} (prefill-decode mixing)}
\centering
\resizebox{\columnwidth}{!}{%
\begin{tabular}{llcc}
\hline
\textbf{Seq. compression} & \textbf{Model} & \textbf{TPOT (Transformers)} & \textbf{TPOT (vLLM)} \\
\hline
2512$\rightarrow$2010 \textcolor{darkgreen}{$\downarrow$ 20\%} & Qwen2.5-Coder-1.5B &
23.6 $\rightarrow$ 23.2 \textcolor{darkgreen}{$\downarrow$ \hphantom{0}2\%} &
\hphantom{0}5.3 $\rightarrow$ \hphantom{0}5.3 \textcolor{darkgreen}{$\downarrow$ \hphantom{0}0\%} \\
 & Qwen2.5-Coder-7B &
27.2 $\rightarrow$ 25.4 \textcolor{darkgreen}{$\downarrow$ \hphantom{0}7\%} &
11.7 $\rightarrow$ 11.7\textcolor{darkgreen}{$\downarrow$ \hphantom{0}0\%} \\
 & Qwen2.5-Coder-14B &
58.5 $\rightarrow$ 52.8 \textcolor{darkgreen}{$\downarrow$ 10\%} &
22.3 $\rightarrow$ 21.8 \textcolor{darkgreen}{$\downarrow$ \hphantom{0}2\%} \\
\hline
2000$\rightarrow$1510 \textcolor{darkgreen}{$\downarrow$ 24\%} & Qwen2.5-Coder-1.5B &
23.5 $\rightarrow$ 23.1 \textcolor{darkgreen}{$\downarrow$ \hphantom{0}2\%} &
\hphantom{0}5.3 $\rightarrow$ \hphantom{0}5.6 {$\uparrow$ \hphantom{0}5\%} \\
 & Qwen2.5-Coder-7B &
25.1 $\rightarrow$ 24.1 \textcolor{darkgreen}{$\downarrow$ \hphantom{0}4\%} &
11.7 $\rightarrow$ 11.6 \textcolor{darkgreen}{$\downarrow$ \hphantom{0}1\%} \\
 & Qwen2.5-Coder-14B &
52.7 $\rightarrow$ 47.0 \textcolor{darkgreen}{$\downarrow$ 11\%} &
21.8 $\rightarrow$ 21.8 \textcolor{darkgreen}{$\downarrow$ \hphantom{0}0\%} \\
\hline
1500$\rightarrow$1010 \textcolor{darkgreen}{$\downarrow$ 33\%} & Qwen2.5-Coder-1.5B &
23.2 $\rightarrow$ 23.5 {$\uparrow$ \hphantom{0}1\%} &
\hphantom{0}5.4 $\rightarrow$ \hphantom{0}6.3 {$\uparrow$ 16\%} \\
 & Qwen2.5-Coder-7B &
24.1 $\rightarrow$ 24.1 \textcolor{darkgreen}{$\downarrow$ \hphantom{0}0\%} &
11.6 $\rightarrow$ 11.5 \textcolor{darkgreen}{$\downarrow$ \hphantom{0}1\%} \\
 & Qwen2.5-Coder-14B &
46.8 $\rightarrow$ 41.1 \textcolor{darkgreen}{$\downarrow$ 12\%} &
21.7 $\rightarrow$ 21.4 \textcolor{darkgreen}{$\downarrow$ \hphantom{0}1\%} \\
\hline
\end{tabular}
}
\label{tab:tpot_transformers_vllm_combined}
\end{table}

Reducing prompt length primarily improves TTFT. In colocated engines, it often yields limited gains on decode-side TPOT, which remains dominated by iterative decode dynamics and batching.
Under disaggregation, the effect becomes more predictable: shorter contexts directly reduce prefill latency and lower the number of GPUs required to handle the same load while leaving decode behavior isolated, allowing clearer SLA tuning for each phase. 


\begin{table}
\setlength{\abovecaptionskip}{5pt}
\caption{Effect of context compression on TTFT on an NVIDIA A100, comparing disaggregated prefill-decode serving (\textnormal{transformers}) against prefill-decode mixing (\textnormal{vLLM}). Under disaggregation, context compression directly leads to a similar decrease in TTFT, enabling an earlier start of response generation. Under prefill-decode mixing, gains are smaller because decode-side workload dominates end-to-end latency.
}
\resizebox{\columnwidth}{!}{%
\begin{tabular}{llccc}
\hline
\textbf{Seq. compression} & \textbf{Model} & \textbf{TFLOPs} & \textbf{TTFT (Transformers)} & \textbf{TTFT (vLLM)} \\
\hline
$2512\rightarrow2010$ \textcolor{darkgreen}{$\downarrow 20\%$} & Qwen2.5-Coder-1.5B & $\hphantom{0}30.9\rightarrow \hphantom{0}24.8$ \textcolor{darkgreen}{$\downarrow 20\%$} & $198.2\rightarrow 156.6$ \textcolor{darkgreen}{$\downarrow 21\%$} & $\hphantom{0}74.7\rightarrow \hphantom{0}68.2$ \textcolor{darkgreen}{$\downarrow \hphantom{0}9\%$} \\
 & Qwen2.5-Coder-7B & $141.5\rightarrow113.8$ \textcolor{darkgreen}{$\downarrow 20\%$} & $668.6\rightarrow 541.1$ \textcolor{darkgreen}{$\downarrow 19\%$} & $\hphantom{}198.3\rightarrow \hphantom{}166.5$ \textcolor{darkgreen}{$\downarrow 16\%$} \\
 & Qwen2.5-Coder-14B & $279.9\rightarrow225.1$ \textcolor{darkgreen}{$\downarrow 20\%$} & $822.8\rightarrow 661.3$ \textcolor{darkgreen}{$\downarrow 20\%$} & $\hphantom{}349.8\rightarrow \hphantom{}291.7$ \textcolor{darkgreen}{$\downarrow 17\%$} \\
\hline
$2000\rightarrow1510$ \textcolor{darkgreen}{$\downarrow 25\%$} & Qwen2.5-Coder-1.5B & $\hphantom{0}24.7\rightarrow\hphantom{0}18.7$ \textcolor{darkgreen}{$\downarrow 24\%$} & $157.4\rightarrow 113.4$ \textcolor{darkgreen}{$\downarrow 28\%$} & $\hphantom{0}65.3\rightarrow \hphantom{0}58.4$ \textcolor{darkgreen}{$\downarrow 11\%$} \\
 & Qwen2.5-Coder-7B & $113.2\rightarrow\hphantom{0}85.5$ \textcolor{darkgreen}{$\downarrow 24\%$} & $540.0\rightarrow 406.8$ \textcolor{darkgreen}{$\downarrow 25\%$} & $\hphantom{}179.0\rightarrow \hphantom{}134.0$ \textcolor{darkgreen}{$\downarrow 25\%$} \\
 & Qwen2.5-Coder-14B & $224.0\rightarrow169.1$ \textcolor{darkgreen}{$\downarrow 25\%$} & $662.2\rightarrow 496.3$ \textcolor{darkgreen}{$\downarrow 25\%$} & $\hphantom{}291.5\rightarrow \hphantom{}232.9$ \textcolor{darkgreen}{$\downarrow 20\%$} \\
\hline
$1500\rightarrow1010$ \textcolor{darkgreen}{$\downarrow 33\%$} & Qwen2.5-Coder-1.5B & $\hphantom{0}18.5\rightarrow\hphantom{0}12.5$ \textcolor{darkgreen}{$\downarrow 32\%$} & $112.2\rightarrow \hphantom{0}69.7$ \textcolor{darkgreen}{$\downarrow 38\%$} & $\hphantom{0}58.9\rightarrow \hphantom{0}50.3$ \textcolor{darkgreen}{$\downarrow 15\%$} \\
 & Qwen2.5-Coder-7B & $\hphantom{0}84.9\rightarrow\hphantom{0}57.2$ \textcolor{darkgreen}{$\downarrow 33\%$} & $406.4\rightarrow 282.2$ \textcolor{darkgreen}{$\downarrow 31\%$} & $\hphantom{}138.0\rightarrow \hphantom{}112.4$ \textcolor{darkgreen}{$\downarrow 19\%$} \\
 & Qwen2.5-Coder-14B & $168.0\rightarrow113.2$ \textcolor{darkgreen}{$\downarrow 33\%$} & $495.6\rightarrow 339.6$ \textcolor{darkgreen}{$\downarrow 31\%$} & $\hphantom{}238.2\rightarrow \hphantom{}174.6$ \textcolor{darkgreen}{$\downarrow 27\%$} \\
\hline
\end{tabular}
}
\label{tab:dissagr_cmp_prefill_decode_full}
\end{table}
~                                                                                                                                                                                                                   
~                                                                                                                                                                                                                   
~

\subsection{Offline Indexing Cost}
\label{sec:rag_build_cost}

Since CoRoVA precomputes compressed context representations during indexing, its offline cost depends on how quickly we can (i) chunk a repository and (ii) run Encoder + Projector over the resulting chunks.
To quantify this overhead, we sample GitHub repositories from The Stack dataset across multiple repository-size quantiles and measure end-to-end time to build the retrieval database: chunking the codebase, encoding each chunk, and applying the projector to obtain the stored compressed representations.

Unless stated otherwise, chunking is performed using a single CPU process, while encoding and projection run on a single A100 GPU.
We do not implement parallelism across these stages, making the reported numbers conservative upper bounds. In practice, indexing can be parallelized across files and performed incrementally as repositories change.

Results are shown in Table~\ref{tab:rag_database_build}.
We find that switching between a 2-layer and 3-layer projector has a negligible impact on database build time, so we report the 3-layer projector configuration throughout.
In contrast, encoder choice substantially affects indexing throughput: the UniXCoder-based pipeline can be up to $10\times$ faster than Qwen3-Embedding in our setup.
Overall, these results indicate that CoRoVA’s offline indexing overhead is practical for IDE workflows and can be further reduced with standard parallel and incremental indexing strategies.

\begin{table}[t!]
\setlength{\abovecaptionskip}{5pt}
\caption{Time to build RAG database (chunking, computing encoder output, and projection). \textnormal{Qwen+3L} = Qwen3-Embedding encoder + 3-layer projector (with $2\times$ expansion), \textnormal{UnxC+3L} = UniXCoder encoder + 3-layer projector.}
\centering
\resizebox{\columnwidth}{!}{%
\begin{tabular}{l c c c c c c}
\hline
\textbf{Repository} &  \textbf{Lines} & \textbf{Files} & \textbf{Chunks} & \textbf{Chunking} & 
\textbf{Qwen+3L} & \textbf{UnxC+3L} \\
\hline 
numpy/numpy & 117,537 & 266 & 9,953 & 4s & 30s & 3s \\
apache/airflow & 245,463 & 896 & 21,678 & 9s & 54s & 7s \\
Azure/azure-sdk-for-python & 1,077,334 & 2,107 & 94,078 & 45s & 4m 16s & 39s \\ 
hawaii-clean-energy-metrics/hcem & 1,083,308 & 51 & 98,858 & 26s & 3m 40s & 24s \\
ComputationalReflection/stypy & 2,955,552 & 1,028 & 240,714 & 2m 28s & 15m 46s & 2m 22s \\
\hline
 \end{tabular}
 }
\label{tab:rag_database_build}
\end{table}

\section{Conclusion}
\label{sec:conclusion}
Retrieval-augmented code completion can substantially improve prediction quality, but the additional retrieved tokens inflate the prefill cost and degrade TTFT, undermining usability in interactive IDE settings. In this work, we introduced CoRoVA, a novel vector-augmented code-completion pipeline that compresses retrieved cross-file context into a small set of continuous vectors via embedding projection, significantly improving code completion quality while preserving low-latency responsiveness. Compared to full RAG, the proposed approach results in 20-38\% better prompt processing speed and latency metrics critical for code completion applications, while maintaining comparable generation quality.

To the best of our knowledge, our work is the first among LLaVA-like approaches to successfully apply compression to code generation models, to explore the addition of semantically rich code modalities, to utilize base models instead of instruction-tuned models, and to apply reinforcement learning to train the projection for downstream code-completion tasks.

CoRoVA is trained by \textit{updating only a lightweight projector module}, keeping both the \textit{encoder and the code LLM frozen}, and using a composite objective that combines cross-entropy with a REINFORCE-style term that directly targets developer-relevant metrics (ES and EM) and a novel cosine alignment loss to prevent representational collapse after projection.
Moreover, we demonstrated that all previously proposed methods for training a projector \emph{fail on code completion tasks}, and designed a complex loss function that \emph{consistently improves the target metrics}.

Future work could investigate more compute-intensive RL optimization strategies (e.g., GRPO~\cite{shao2024deepseekmath}) to further improve alignment with EM/ES metrics, and revisit structural signals as stronger graph-based encoders are developed.



\bibliographystyle{ACM-Reference-Format}
\balance
\bibliography{bibliography}

\end{document}